\documentclass{article}
\usepackage[preprint]{neurips_2026}  % arXiv preprint: de-anonymized, authors shown
\usepackage[utf8]{inputenc}
\usepackage[T1]{fontenc}
\usepackage{hyperref}
\usepackage{url}
\usepackage{amsfonts}
\usepackage{nicefrac}
\usepackage{microtype}
\usepackage{graphicx}
\usepackage{wrapfig}
\usepackage{tcolorbox}
\tcbuselibrary{breakable, listings, skins}
\usepackage{xcolor}
\usepackage{listings}
\usepackage{booktabs}
\definecolor{promptvar}{HTML}{C04020}
\definecolor{promptbg}{HTML}{F7F6F2}
\definecolor{promptframe}{HTML}{6B6B6B}

\lstdefinestyle{prompttpl}{
    basicstyle=\ttfamily\footnotesize,
    breaklines=true,
    breakatwhitespace=false,
    breakindent=0pt,
    columns=fullflexible,
    keepspaces=true,
    showstringspaces=false,
    upquote=true,
    extendedchars=true,
    inputencoding=utf8,
    moredelim=*[s][{\itshape\color{promptvar}}]{<<}{>>},
    literate=%
        {—}{{---}}1
        {–}{{--}}1
        {→}{{$\rightarrow$}}1
        {←}{{$\leftarrow$}}1
        {≥}{{$\geq$}}1
        {≤}{{$\leq$}}1
        {≠}{{$\neq$}}1
        {…}{{\ldots}}1
        {±}{{$\pm$}}1
        {×}{{$\times$}}1
        {÷}{{$\div$}}1,
}

\newtcblisting{prompttemplate}[2][]{
    listing only,
    listing options={style=prompttpl},
    title={#2},
    colback=white,
    colframe=promptframe,
    coltitle=white,
    colbacktitle=promptframe,
    fonttitle=\bfseries,
    boxrule=0.4pt,
    enhanced,
    breakable,
    sharp corners=southwest,
    left=4pt, right=4pt, top=4pt, bottom=4pt,
    #1
}

\newcommand{\methodname}{\textsc{AutoMem}}
\newcommand{\qwen}{\texttt{Qwen2.5-32B-Instruct}}
\newcommand{\pms}[1]{$\pm$ #1}
\renewcommand{\acksection}{\section*{Acknowledgments}}

\title{\methodname: Automated Learning of Memory as a Cognitive Skill}
\author{%
  Shengguang Wu \quad Hao Zhu \quad Yuhui Zhang \quad Xiaohan Wang \quad Serena Yeung-Levy \\[2pt]
  Stanford University \\[2pt]
  ~ \\[2pt]
  \url{https://autolearnmem.github.io/}
}

\begin{document}
\maketitle

\begin{abstract}
Memory expertise is a learned skill: knowing what to encode, when to retrieve, and how to organize knowledge---a capacity known in cognitive science as \textit{metamemory}. We bring this perspective to LLMs by treating memory management as a trainable skill.
We promote file-system operations to first-class memory actions alongside task actions, letting the model itself decide how to manage its memory.
This memory skill improves along two axes: the \textbf{\textit{structure}} that supports it (prompts, file schemas, action vocabulary), and the \textbf{\textit{proficiency}} of the model exercising it.
Both axes resist manual optimization: episodes in long-horizon tasks run for thousands of steps, and a single memory mistake can hide long before it surfaces, making human review of full trajectories impractical.
We introduce \methodname{}, a framework that \textbf{automates} both axes.
In the first loop, a strong LLM reviews complete agent trajectories and iteratively revises the memory \textit{structure} that shapes how the agent interacts with its memory files.
In the second loop, the agent's own good memory decisions are identified from many episodes and used as training signal to sharpen the model's memory \textit{proficiency} directly.
Across three procedurally generated long-horizon games (Crafter, MiniHack, and NetHack), optimizing memory alone---without modifying the model's task-action behavior---improved the base agent's performance ${\sim}2\times$--$4\times$, bringing a 32B open-weight model competitive with frontier systems such as Claude Opus 4.5 and Gemini 3.1 Pro Thinking.
Our results show that memory management is an independently learnable skill, and a high-leverage objective yielding large gains on long-horizon tasks.
\end{abstract}

\begin{figure}[!h]
    \centering
    \includegraphics[width=\textwidth]{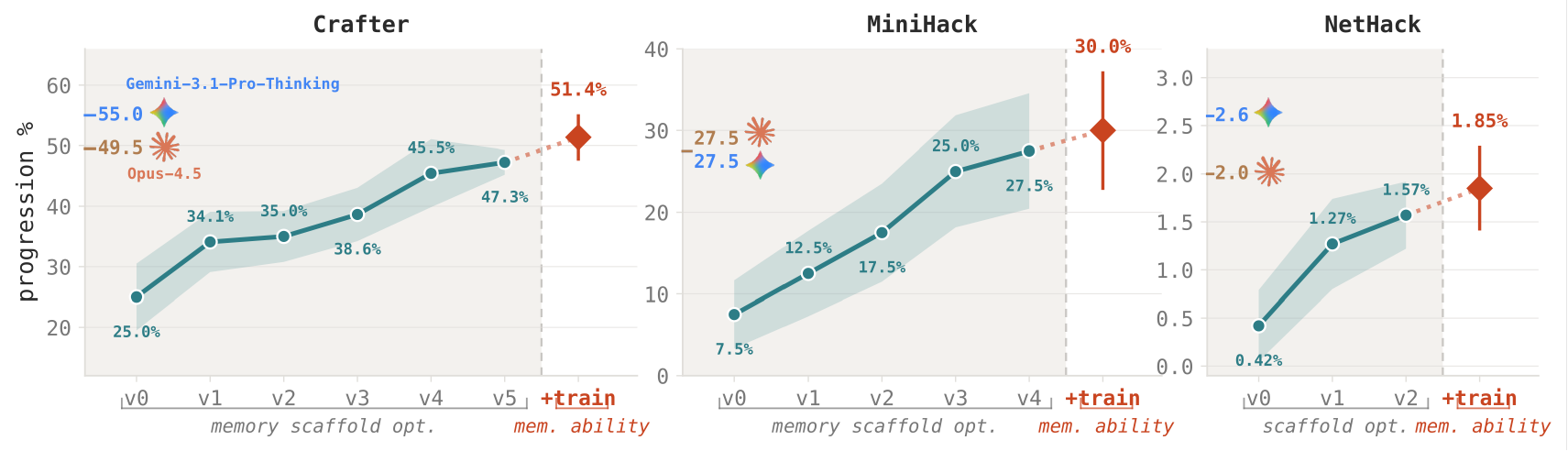}
    \caption{\textbf{Memory skill optimization} with \qwen{}. Starting from a base agent equipped with file-system memory (v0), \methodname{} progressively improves performance through \textit{memory scaffold optimization} (v0--v5/v4/v2), followed by \textit{memory proficiency training} (+train) that yields further gains on top of the optimized scaffold.}
    \label{fig:teaser}
\end{figure}

\section{Introduction}

Humans routinely manage information beyond what can be held in mind at any one moment. Cognitive scientists call this capacity \textbf{\textit{metamemory}}: the learned skill of deciding what is worth remembering, when to retrieve it, and how to organize what is known~\citep{flavell1979metacognition,nelson1990metamemory}. Metamemory develops with practice, and skilled use of external aids---notes, indices, files---is part of how people extend cognition beyond working memory~\citep{clark1998extended}.

LLMs face an analogous bottleneck. Their context window plays the role of working memory, \emph{i.e.,} a fixed-size buffer that bounds what the model can attend to at once. 
Long-horizon tasks routinely exceed this capacity, and external memory has been explored in various forms, including retrieval databases, vector stores, scratchpads, and summary buffers~\citep{lewis2020rag,packer2023memgpt,park2023generative,xu2025amem,sumers2023cognitive,zhang2024survey,zhong2024memorybank}. These approaches typically treat memory as an architectural module: a fixed mechanism designed into the system. 
We take a different view, one inspired by \textit{metamemory}: \textbf{memory management is an active, trainable skill}, and the model itself decides what to store, what to look up, and how to structure its records.

Concretely, we promote file-system operations (read, write, search, append, create) as \textit{first-class memory actions} in the model's action space, on equal footing with the actions it uses to act on the world~\citep{yao2022react}. The same forward pass that picks a task action can also select memory file operations (\emph{e.g.,} \texttt{<|APPEND|>} or \texttt{<|SEARCH|>}). This minimal design gives the model full control over its external memory while keeping behavior cleanly observable: every memory decision is a traceable action in the trajectory.

A learned memory skill improves along two axes. First, there is the \textbf{\textit{structure}} that supports it (the prompts, the file schema, the validation logic, the action vocabulary), which determines what memory operations are available and how the model is guided to use them. 
Second, there is the \textbf{\textit{proficiency}} of the agent exercising the skill---the model's parametric ability to decide well among the available operations. 

Both axes resist manual optimization. A single episode can run for tens of thousands of steps, and the effect of a memory operation (or a mistake) may remain hidden for many steps before it surfaces as a guiding signal or a missed objective. The learning of memory skill in long-horizon tasks is therefore almost intractable for human review.

The key observation behind our approach is that a sufficiently capable LLM---acting as a \textit{meta-LLM}---can review an agent's complete episode (spanning thousands of steps) and identify where memory decisions went wrong, much as a code reviewer would read a full execution log. This makes it possible to \textbf{automate} both axes of memory improvement.
We introduce \methodname{} (Figure~\ref{fig:method}), which does so through two sequential outer loops that operate on a shared inner-loop agent using a file system as its memory.

In the first loop (\textbf{\textit{structure}}), a meta-LLM reads complete episode traces, diagnoses failure patterns in the agent's memory use, and iteratively revises the agent scaffold: the code, prompts, and memory file schema that shape how the agent interacts with memory and acts on the world.

In the second loop (\textbf{\textit{proficiency}}), the agent's own memory decisions from many episodes are reviewed and the ones worth reinforcing are selected---again by a meta-LLM---as supervised training data for a dedicated memory model. The same meta-LLM also orchestrates the finetuning configuration that absorbs this data.
Because we treat memory as a separable skill, we finetune only a dedicated memory model (\textit{memory specialist}) while the model that commits world actions remains unmodified, thus sharpening memory proficiency without risking the agent's existing task competence.

A common principle connects both loops: \textbf{long-horizon task improvement can decompose into trajectory-level review and targeted revision}---a workflow that a strong meta-LLM can execute autonomously where human review of complete traces (up to $10^5$ steps) is impractical. The first loop applies this principle to revise code; the second applies it to curate training data and orchestrate the training itself.

\begin{figure}[t]
    \centering
    \includegraphics[width=\textwidth]{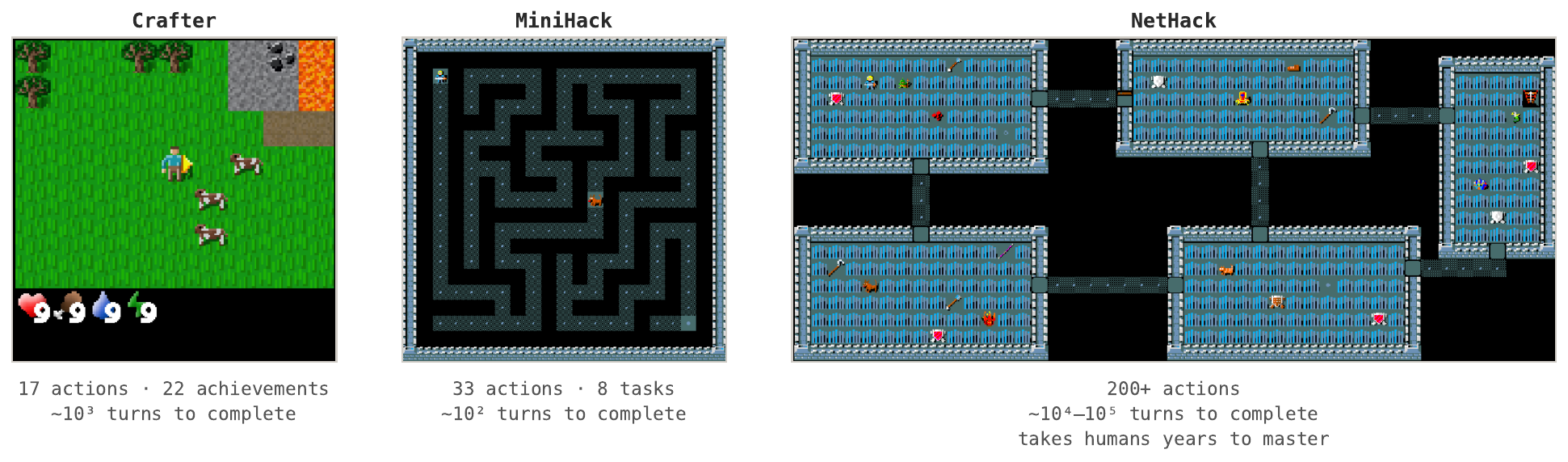}
    \caption{\textbf{Long-horizon game environments for evaluating memory skills.} All three environments are stochastic worlds, making each episode unique and minimizing the influence of pretraining knowledge. \textbf{Crafter} is an open-world survival game with crafting, combat, and resource management. \textbf{MiniHack} presents focused puzzle, navigation and combat tasks within the NetHack engine. \textbf{NetHack} is among the most complex games: episodes span $10^4$--$10^5$ turns with a vast exploration space, taking human players typically years to master.}
    \label{fig:envs}
\end{figure}

We evaluate on procedurally generated long-horizon games, which are well suited for studying memory skill: episodes are long enough that context-window management alone cannot sustain performance; worlds are regenerated each episode, so pretraining knowledge transfers poorly~\citep{paglieri2024balrog}; and success demands the kind of records humans naturally keep---such as maps, inventories, encounter logs, strategy notes.
We choose three environments spanning span a range of complexity (Figure~\ref{fig:envs}): \textbf{Crafter}, an open-world survival game with crafting and resource management~\citep{hafner2021benchmarking}; \textbf{MiniHack}, a suite of focused puzzle, navigation and combat tasks~\citep{samvelyan2021minihack}; and \textbf{NetHack}, a roguelike whose $10^4$--$10^5$ step episodes take human players years to master~\citep{kuttler2020nethack}.

Using \qwen{} as the base model, optimizing memory alone---without modifying the model's task-action weights---the full \methodname{} framework yields ${\sim}2\times$--$4\times$ gains over the base agent (Table~\ref{tab:main}, Figure~\ref{fig:teaser}).
The optimized 32B agent outperforms \texttt{Qwen2.5-72B-Instruct} on all three games by a wide margin, indicating that \textbf{well-managed memory is higher-leverage than model scale} on these tasks.
It also reaches the performance level of frontier proprietary systems such as Claude Opus 4.5 and Gemini 3.1 Pro Thinking---showing that the gap between open-weight and frontier models on long-horizon tasks can be substantially closed by targeting memory as the optimization objective.

\paragraph{Contributions.}
\emph{(i)} We formulate memory management as an independently learnable skill for LLM agents, instantiated through file-system operations that sit in the same action space as task actions, giving the model full, observable control over what to encode, when to retrieve, and how to organize its memory.
\emph{(ii)} We introduce \methodname{}, a framework that automates memory skill improvement along two complementary axes: scaffold revision that iterates on the agent's memory structure, and targeted training of the model's memory proficiency on its own experience. Both loops are driven by meta-LLMs that analyze complete episode traces, making long-horizon optimization feasible where human review of full trajectories is not.
\emph{(iii)} Across three procedurally generated long-horizon games, targeting memory yields ${\sim}2\times$--$4\times$ progression gains for a 32B open-weight model, significantly closing the gap with frontier proprietary systems, and demonstrating that memory is a high-leverage objective for long-horizon tasks.

\section{\methodname{}} \label{sec:method}

\begin{figure}[t]
    \centering
    \includegraphics[width=\textwidth]{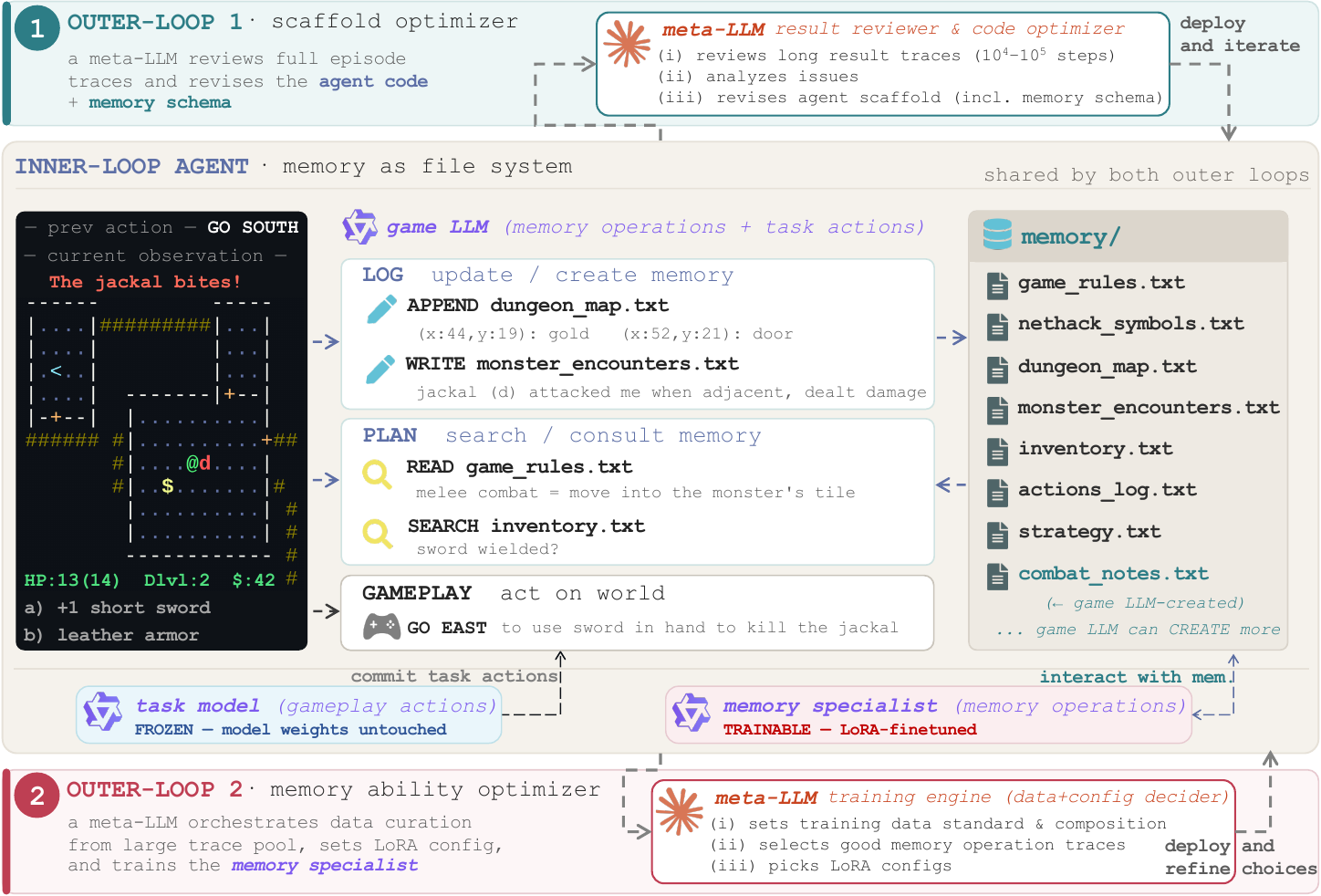}
    \caption{\textbf{Overview of \methodname{}.} Two automated outer loops optimize a shared inner-loop agent that uses the file system as its memory. \textbf{Outer-loop \#1} (top): a meta-LLM reviews full episode traces and iteratively revises the agent scaffold. \textbf{Outer-loop \#2} (bottom): a meta-LLM \emph{training engine} jointly orchestrates data curation and finetuning configuration to train a dedicated \emph{memory specialist} that handles memory operations, while the \emph{task model} (frozen, unmodified) commits task actions. The two loops are complementary: loop~\#1 produces an optimized scaffold within which loop~\#2 trains the model to interact with its memory more effectively.}
    \label{fig:method}
\end{figure}

Memory skill, as formulated above, improves along two axes---\textit{structure} and \textit{proficiency}---both of which require automation to optimize on long-horizon tasks. 
\methodname{} provides this automation through two sequential outer loops that operate on a shared inner-loop agent (Figure~\ref{fig:method}). The first loop pushes the agent scaffold as far as code revision can take it; the second trains the model's memory ability beyond the ceiling of any fixed scaffold.

\subsection{Inner-loop agent: memory as file system}
\label{sec:inner-loop}

The inner loop is a single LLM agent executing one episode of a long-horizon task, equipped with a directory of files on disk that serves as its external memory (gray-shaded area in the middle of Figure~\ref{fig:method}).
At each step the agent runs two routines, each targeting one side of memory management. 
The \textsc{LOG} routine asks ``\textbf{\textit{what is worth recording about what just happened}}'': the agent decides whether and how to record the environment's response to the previous action, \emph{e.g.,} appending to an existing file, creating a new one, or rewriting an entry.
The \textsc{PLAN} routine asks ``\textbf{\textit{what do I need to recall to act now}}'': the agent searches across files, reads specific entries or their tails, and commits the next world action. 

This unified action space is what makes memory a learnable skill rather than a fixed mechanism. 
The file-system substrate gives the model a wide decision space---which files to keep, what to record in each, when to consult them, how to organize what is known---and because every memory decision is a traceable action in the trajectory, the outer loops can observe, evaluate, and optimize it. 

As a result of this shared space, borne out in our experiments (Section~\ref{sec:results}), \textbf{optimizing the memory structure also improves task behavior} (\emph{e.g.,} gameplay actions): better-organized memory reduces redundant exploration and directionless action, even though the optimizer targets the memory scaffold rather than the task strategy directly.

\subsection{Outer-loop 1: optimizing the memory scaffold}
\label{sec:scaffold}

The first outer loop optimizes the \textit{structure} that supports the memory skill. 
The agent scaffold, \emph{i.e.,} the code, prompts, file schema, and action vocabulary that shape how the agent manages memory and acts on the world, is iteratively revised by the meta-LLM.

The optimization signal must be trajectory-level, because the consequences of memory decisions are often delayed in long-horizon tasks. 
A memory mistake at step~50---failing to record a map coordinate, or writing a duplicate entry that buries useful information---may not surface until much later (\emph{e.g.,} step~800), when the agent gets lost or wastes time re-exploring. 
Final-return metrics alone discard the trajectory structure that reveals \emph{where} memory went wrong.

The meta-LLM is therefore given full episode traces (per-step logs, the resulting memory directories, and the agent code itself) and identifies points where the scaffold caused failures. It functions as a code reviewer with a complete execution log in hand, not as a scalar reward signal. 
For example, reviewing NetHack traces, the meta-LLM identified that an unbounded map file was accumulating thousands of duplicate coordinate entries, burying useful information; it responded by introducing a clean coordinate-keyed map deduplication format (Figure~\ref{fig:scaffold-evolution}) that sharply shrank the map the agent must carry.

Each iteration is gated on measured improvement: the rewritten agent plays the same fixed seeds as the previous version, and the revision is kept only if average progression improves. Details about the gate and retry mechanics are in Appendix~\ref{app:method-impl}. 
At rough ``convergence'' (in practice, after \emph{e.g.,} 2--5 iterations as shown in Figure~\ref{fig:teaser}), the scaffold has absorbed what code revision can express. 
Concrete revisions the optimizer produced across iterations are discussed in Section~\ref{sec:results} and listed in Appendix~\ref{app:scaffold-evolution}. Figure~\ref{fig:scaffold-evolution} illustrates how the memory file schema evolved.
Given an agent scaffold that is already lean and well-structured, the model's parametric ability to navigate its memory becomes the remaining bottleneck blocking optimal memory decisions.

\subsection{Outer-loop 2: training memory proficiency} \label{sec:proficiency}

Once the scaffold is optimized, the remaining gap lies in the model's parametric ability to make good memory decisions---the \textit{proficiency} axis.
Where the first loop revised code with the meta-LLM as reviewer, this loop updates model weights with the meta-LLM as a \textit{training engine}---a meta-level optimizer that orchestrates the supervised training process end-to-end.

The memory ability is trained on the model's own experience: the meta-LLM reads the inner-loop agent code together with a pool of episode traces from many random episodes, derives selection criteria from what the agent code requires of the model, and produces supervised training data. 
Every example in the training set is verbatim text the inner-loop model produced during an episode; the meta-LLM's role is to select which responses to reinforce---it thus acts as a \textbf{\textit{filter}} on the model's own behavior rather than as a \textbf{\textit{teacher}} generating new responses.

High-quality data curation is essential, but the training must be \textbf{configured} to absorb it. There are many valid ways to curate good memory traces from the same trace pool (\emph{e.g.,} different selection criteria, sample sizes). Under our LoRA setup---a small adapter finetuned on a modest dataset---a configuration that suits one dataset can underfit or overfit under another, so each data curation calls for its own matching LoRA finetuning configuration to learn effectively. 
The meta-LLM \textit{training engine} therefore orchestrates the data selection logic, the data composition, and the LoRA training configuration as a \textbf{joint decision}, refining all three across iterative trials with concrete eval-trajectory feedback. The point is to find a properly matched (data, configuration) combination so that the finetuning contributes a clean lift of memory capability rather than unwanted disruption.

Because memory is a separable skill, we train it as a distinct target rather than finetuning on full episodes that mix memory operations with action-format outputs.
This separation carries through to inference deployment. The inner loop runs two model instances that share a single conversation history (lower panel of Figure~\ref{fig:method}). The \emph{memory specialist} is the LoRA-finetuned~\citep{hu2022lora} copy and handles the \textsc{LOG} routine together with the memory-consultation portion of the \textsc{PLAN} routine. The \emph{gameplay model} is the unmodified base and commits the world action. 
After the specialist's last memory operation, a short handoff passes the conversation to the gameplay model, which can still issue further memory reads before committing.

Key benefits follow from this split. 
First, \textbf{the training signal stays focused}: the supervised loss targets memory-operation behavior without being diluted by action-format examples, so each gradient step addresses memory proficiency directly. 
Second, \textbf{the base model's competence at producing well-formatted world actions is fully preserved}, since the gameplay model's weights are never touched by finetuning. 
In practice, this means the memory-proficiency gain stacks cleanly on top of the scaffold gain (a further ${\sim}9$--18\% relative gain, see Section~\ref{sec:results}) rather than trading off against it.

\paragraph{Two loops, one objective.} Combining both loops together, \methodname{} instantiates a coherent process for \textbf{learning memory as a skill}, one that mirrors the familiar structure of machine learning.
Each loop has \textit{parameters} $\theta$ to optimize and an \textit{update signal} $\nabla L$ derived from the meta-LLM's trajectory analysis---but in the first loop $\theta$ is the agent scaffold and $\nabla L$ is a code revision, while in the second, $\theta$ is the dedicated memory model weights and $\nabla L$ is a properly-configured supervised training step (curated data together with a matched LoRA setup) drawn from the agent's own behavior.
The scaffold sets the structural ceiling for what memory operations are possible, and proficiency training then pushes the model toward that ceiling.

\section{Experiments}

\begin{table}[t]
    \centering
    \caption{\textbf{Performance on long-horizon games from BALROG.} \methodname{} is built on \qwen{}. Frontier numbers are taken from the BALROG leaderboard~\citep{paglieri2024balrog}; baseline rows for \qwen{} use the same BALROG harness with the listed context management method. \textit{Scaffold opt.} refers to the agent produced by Outer-loop \#1 at convergence (Crafter at \texttt{v5}, MiniHack at \texttt{v4}, NetHack at \texttt{v2}). \textit{+ memory training} adds Outer-loop \#2's training engine on top. All values are progression rate (\%), reported as mean $\pm$ standard error.}
    \label{tab:main}
    \small
    \renewcommand{\arraystretch}{1.0}
    \begin{tabular*}{\linewidth}{@{}p{0.52\linewidth}@{\extracolsep{\fill}}ccc@{}}
    \toprule
    Agent & Crafter (\%) & MiniHack (\%) & NetHack (\%) \\
    \midrule
    \multicolumn{4}{@{}l}{\textit{Frontier proprietary (BALROG leaderboard)}} \\[2pt]
    Gemini-3-Pro                      & 57.3 \pms{4.4} & 40.0 \pms{7.7} & 6.8 \pms{3.2} \\
    Gemini-3.1-Pro-Thinking           & 55.0 \pms{6.4} & 27.5 \pms{7.1} & 2.6 \pms{0.3} \\
    Claude-Opus-4.5                   & 49.5 \pms{3.1} & 27.5 \pms{7.1} & 2.0 \pms{0.5} \\
    Gemini-2.5-Pro                    & 55.0 \pms{6.0} & 17.5 \pms{6.0} & 1.7 \pms{0.2} \\[2pt]
    \midrule
    \multicolumn{4}{@{}l}{\textit{Open-weight (BALROG leaderboard)}} \\[2pt]
    DeepSeek-R1                       & 36.4 \pms{3.8} & 25.0 \pms{6.8} & 1.4 \pms{0.5} \\
    Qwen2.5-72B-Instruct              & 27.3 \pms{3.6} &  5.0 \pms{3.4} & 0.3 \pms{0.3} \\
    Qwen2.5-7B-Instruct               & 16.4 \pms{3.0} &  0.0 \pms{0.0} & 0.0 \pms{0.0} \\[2pt]
    \midrule
    \multicolumn{4}{@{}l}{\qwen{} \textit{with basic context-management baselines}} \\[2pt]
    \quad sliding window              & 19.55 \pms{3.46} & 2.50 \pms{2.47} & 0.00 \pms{0.00} \\
    \quad + chain-of-thought          & 17.27 \pms{2.71} & 10.00 \pms{4.74} & 0.00 \pms{0.00} \\
    \midrule
    \multicolumn{4}{@{}l}{\qwen{} \textit{with \methodname{} (ours)}} \\[2pt]
    \quad memory-as-file-system, \texttt{v0}      & 25.00 \pms{5.50} &  7.50 \pms{4.16} & 0.42 \pms{0.37} \\
    \quad + scaffold opt.\ (loop \#1)             & 47.27 \pms{2.05} & 27.50 \pms{7.06} & 1.57 \pms{0.35} \\
    \quad + memory training (loop \#2)            & \textbf{51.36} \pms{3.81} & \textbf{30.00} \pms{7.25} & \textbf{1.85} \pms{0.44} \\
    \bottomrule
    \end{tabular*}
\end{table}

\subsection{Setup}
\label{sec:setup}

\paragraph{Environments.} 
We evaluate on three procedurally generated long-horizon games used in BALROG~\citep{paglieri2024balrog} (Figure~\ref{fig:envs}). 
\textbf{Crafter}~\citep{hafner2021benchmarking} is a 2D survival world with 17 actions and 22 achievements covering exploration, crafting, and combat; episodes run up to ${\sim}10^3$ steps. 
\textbf{MiniHack}~\citep{samvelyan2021minihack} is an 8-task suite (mazes, corridors, Boxoban puzzles, quests) built on the NetHack engine, with 33 actions and ${\sim}10^2$ steps per task. 
\textbf{NetHack}~\citep{kuttler2020nethack} is the full NetHack Learning Environment (NLE), with 200+ actions and $10^4$--$10^5$ steps per episode; the dungeon, monsters, and items are regenerated every seed.
The BALROG harness is used as released, with minor configuration changes documented in Appendix~\ref{app:env-config}.

\paragraph{Metric.} 
We use game \emph{progression rate} following BALROG, scaled to $[0, 100]$. 
For Crafter it is the fraction of the 22 achievements obtained in an episode. 
For MiniHack it is the fraction of the 8 task variants completed (binary per task). 
For NetHack it is the dungeon-and-experience-level progression metric introduced by~\citet{paglieri2024balrog}. 
We report mean and standard error over a fixed list of 10 seeds, $[42, 43, \ldots, 51]$, paired with per-environment episode counts: 10 episodes for Crafter, 5 per task for MiniHack ($5 \times 8 = 40$ episodes), and 5 for NetHack, matching the BALROG defaults.

\paragraph{Base model and meta-LLM.} 
The inner-loop model is \qwen{}, served locally with vLLM. 
The scaffold optimizer (Outer-loop~\#1) and the training engine (Outer-loop~\#2) call Claude Opus~4.6 \& Opus~4.7 respectively as the meta-LLM.
In outer-loop~\#2, the \emph{memory specialist} is trained with LoRA; the gameplay model is the unmodified base.  
Hyperparameters, prompt templates and more implementation details are in Appendix~\ref{app:setup}.

\paragraph{Baselines.} Two groups. (i) BALROG leaderboard: frontier proprietary models (Gemini-3-Pro, Gemini-3.1-Pro-Thinking, Claude-Opus-4.5, Gemini-2.5-Pro), frontier open-weight reasoning models (DeepSeek-R1, 671B), and same-family open weights at two scales (Qwen2.5-72B-Instruct, Qwen2.5-7B-Instruct). (ii) \qwen{} under the BALROG harness with basic context-management methods: a sliding window of $16$-step recent observations with/without chain-of-thought~\citep{wei2022chain}.

\subsection{Results} \label{sec:results}

\begin{figure}[b]
  \centering
  \includegraphics[width=\linewidth]{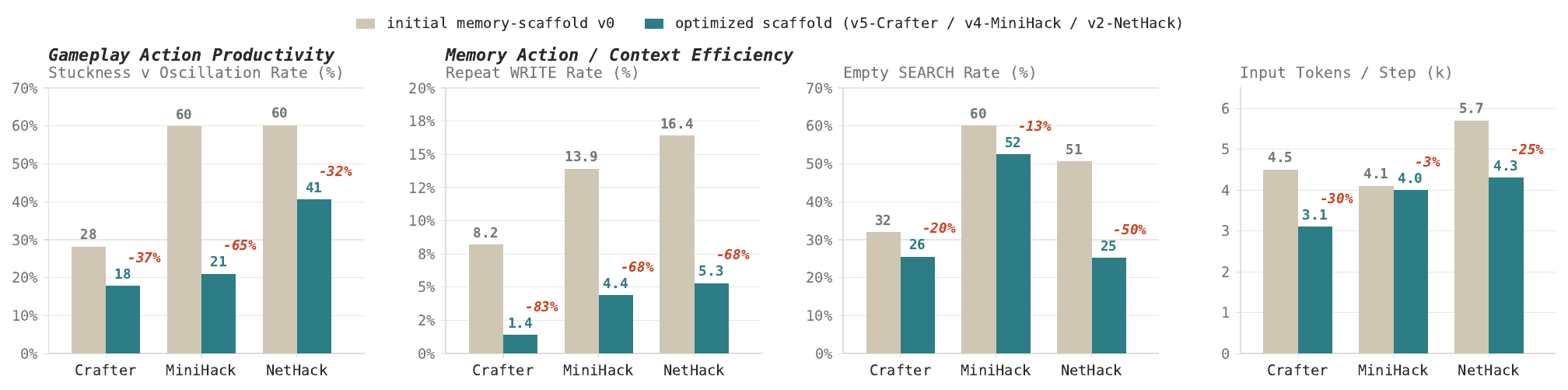}
  \caption{\textbf{Effect of scaffold optimization on gameplay and memory behavior} (v0~$\to$~v5 for Crafter, v0~$\to$~v4 for MiniHack, v0~$\to$~v2 for NetHack). \textit{Left:} the unproductive game action rate (fraction of steps that are either stuck or oscillating) drops 32--65\% across all three environments. \textit{Right three panels:} \textbf{memory operations become more efficient and better targeted}---redundant writes drop sharply ($-$68 to $-$83\%), the empty-search rate (memory \texttt{SEARCH}es returning nothing) falls ($-$13 to $-$50\%), and per-step input context shrinks ($-$3 to $-$30\%) as leaner memory compresses what the model must attend to. All values are v0 vs.\ final scaffold version; lower is better in every panel.}
  \label{fig:scaffold-stats}
\end{figure}

\paragraph{Memory management is a high-leverage axis on its own.}
The model weights are untouched throughout scaffold optimization---only the agent code, prompts, and file schema change---yet progression \textbf{roughly doubles or more than triples on every environment}: Crafter 25.0\%$\to$47.27\% ($\times$1.89), MiniHack 7.5\%$\to$27.5\% ($\times$3.67), NetHack 0.42\%$\to$1.57\% ($\times$3.74) (Table~\ref{tab:main}, Figure~\ref{fig:teaser}).

To calibrate this: the scaffolded 32B agent outperforms Qwen2.5-72B-Instruct (under the BALROG harness) by a wide margin, and surpasses the same base model under basic sliding-window context strategy by an even wider one, indicating that well-structured external memory management is a \textbf{higher-leverage axis than model scale} or context management on these long-horizon tasks.

\paragraph{Training the memory specialist adds complementary lift on top of the optimized scaffold.}
On top of the optimized scaffold (\texttt{v5} for Crafter, \texttt{v4} for MiniHack, \texttt{v2} for NetHack), the memory-proficiency training loop lifts progression to 51.36\% on Crafter (+4.09), 30.0\% on MiniHack (+2.5), and 1.85\% on NetHack (+0.28), comparable in magnitude to one or two scaffold iterations (Table~\ref{tab:main}).

The two optimization axes thus work in sync: structure-revision (outer-loop 1) pushes the ceiling of prompts and agent schema, and proficiency-training (outer-loop 2) addresses the remaining model-capacity gap.
Together they bring the 32B open-weight model to the performance level of frontier proprietary systems on these tasks, comparable to Claude-Opus-4.5 (Crafter/MiniHack/NetHack: 49.5/27.5/2.0) and within a few points of Gemini-3.1-Pro-Thinking (55.0/27.5/2.6).

\paragraph{Task decisions and memory operations are simultaneously improved in the shared action space.}
Because memory operations and task actions share the same action space, both are observable in full episode traces, and the meta-LLM's trajectory-level review surfaces failures on both sides. 
We measure four behavioral indicators to quantify the effect of scaffold optimization (Figure~\ref{fig:scaffold-stats}).

On the task action (\emph{i.e.,} gameplay) side, we track the \textbf{\textit{unproductive action rate}}: the fraction of steps where the agent is either \textbf{stuck} (observation unchanged from the previous step---the action had no effect) or \textbf{oscillating} ($\geq$3 direction reversals in the last 10 movements---pacing back and forth without spatial memory of where it has been).
This combined rate drops 32--65\% across all three environments. This reduction recovers steps that the base agent spent repeating ineffective actions; each recovered step is an opportunity to explore, gather, or craft that was previously wasted.

On the memory side, the optimized scaffold makes memory operations substantially more efficient and better targeted.
\textit{Redundant memory-file writes} drop sharply ($-$68 to $-$83\%) and the \textit{empty-search rate}---memory \texttt{SEARCH}es that return nothing---falls ($-$13 to $-$50\%), so the agent both \textbf{writes less duplicate content} and \textbf{retrieves more precisely}. Especially on Crafter and NetHack, the leaner memory also compresses the \textit{input context} per step ($-$25 to $-$30\% tokens).
This precisely echoes the purpose of external memory, \emph{i.e.,} \textbf{compressing the information} the model needs to attend to. 
The optimized scaffold leads to the emergence of such effective compression because the meta-LLM's trajectory review surfaces the consequences of verbose or redundant memory on the actual task environments---a key benefit of memory operations and task decisions sitting in the shared action space.
\begin{wrapfigure}{r}{0.6\textwidth}
  \centering
  \includegraphics[width=\linewidth]{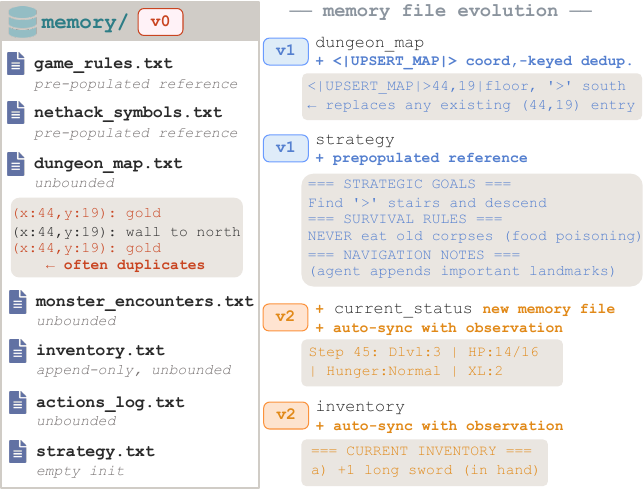}
  \caption{\textbf{Example memory file evolution (NetHack).} The base schema (v0) logs to an unbounded, append-only \texttt{dungeon\_map.txt} that accumulates duplicate coordinate entries. Scaffold optimization replaces this with a coordinate-keyed \texttt{<|UPSERT\_MAP|>} deduplication format, adds auto-synced \texttt{inventory} and \texttt{status} files maintained from observation, and pre-populates a \texttt{strategy} reference.}
  \label{fig:scaffold-evolution}
\end{wrapfigure}

Figure~\ref{fig:scaffold-evolution} illustrates some concrete memory schema changes that drive these numbers, using NetHack as an example. More details on per-iteration changes for all three environments are listed in Appendix~\ref{app:scaffold-evolution}.
The base scaffold (v0) uses an append-only \texttt{dungeon\_map.txt} that grows without bound, frequently accumulating duplicate coordinate entries (the same tile logged multiple times as the agent revisits it).
The optimizer introduces a dedicated \texttt{<|UPSERT\_MAP|>} operation that replaces append-based map logging with coordinate-keyed deduplication, so any new observation of tile $(x, y)$ overwrites the previous entry rather than appending alongside it. 
The evolved schema also adds auto-synced \texttt{inventory} and \texttt{status} files that the scaffold updates programmatically from the observation, removing the need for the model to manually \texttt{READ} and reconcile its own records, and a pre-populated \texttt{strategy} reference that encodes the game's primary objective (find stairs and descend) so the model does not waste early-episode operations rediscovering the goal.
Together these changes sharply shrink the memory the agent must carry: its per-step growth falls from 138 to 6 characters, a 95\% reduction.
\paragraph{Training internalizes a consult-before-write memory discipline.}
\begin{wraptable}{r}{0.44\textwidth}
\vspace{-1.2\baselineskip}
\centering
\small
\caption{LOG-phase memory \texttt{writes} per \texttt{SEARCH} on the \emph{evolved} scaffold with base model vs. \textbf{+ trained} memory specialist (lower $=$ more retrieval before writing). The trained specialist checks memory before writing in every environment.}
\begin{tabular}{lcc}
\toprule
 & \textbf{Base} & \textbf{+ Trained} \\
\midrule
Crafter  & 0.84 & 0.39 \textcolor{gray}{\scriptsize($-$54\%)} \\
MiniHack & 2.89 & 0.82 \textcolor{gray}{\scriptsize($-$72\%)} \\
NetHack  & 4.66 & 1.31 \textcolor{gray}{\scriptsize($-$72\%)} \\
\bottomrule
\end{tabular}
\label{tab:training-search}
\vspace{-1.0\baselineskip}
\end{wraptable}
Beyond the aggregate progression lift, the trained memory specialist exhibits a consistent shift toward \textbf{consulting memory before modifying it} (Table~\ref{tab:training-search}).
In the LOG phase, the ratio of memory \texttt{writes} to \texttt{SEARCH}es falls in every environment: Crafter $0.84\to0.39$ ($-$54\%), MiniHack $2.89\to0.82$ ($-$72\%), and NetHack $4.66\to1.31$ ($-$72\%). That is, the specialist searches existing files more before appending new content, rather than writing blindly.
This is precisely the consult-before-write pattern the optimized scaffold encourages via prompting, now \textbf{internalized} into the dedicated memory model weights.

\begin{figure}[p]
  \centering
  \includegraphics[width=\linewidth]{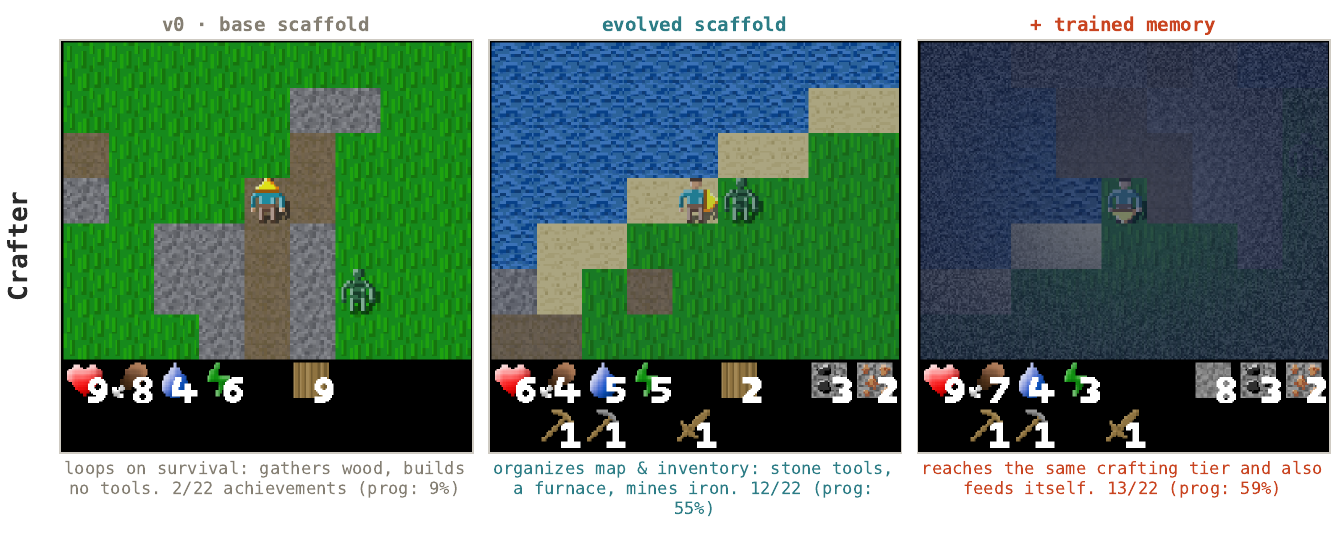}\\[3pt]
  \includegraphics[width=\linewidth]{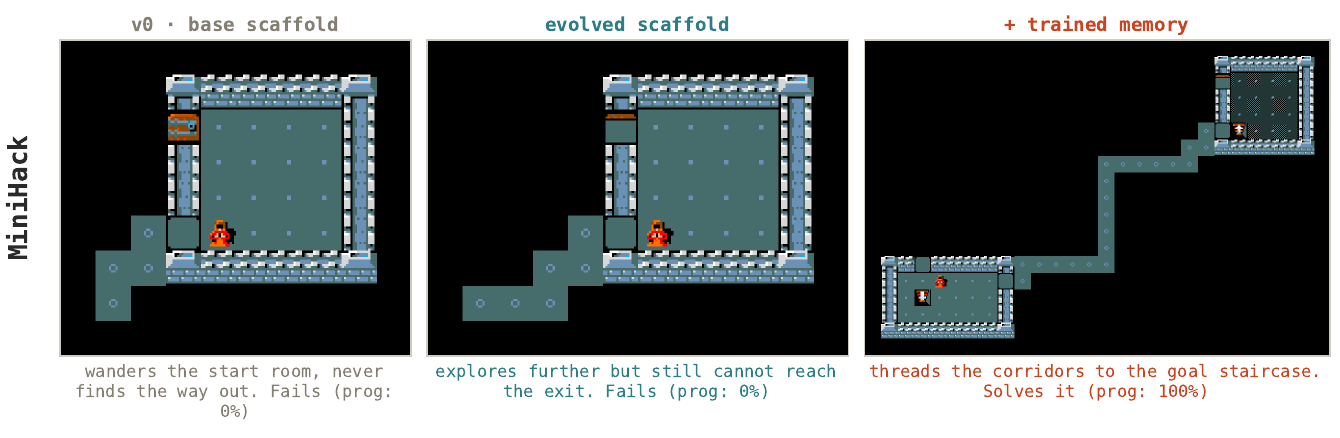}\\[3pt]
  \includegraphics[width=\linewidth]{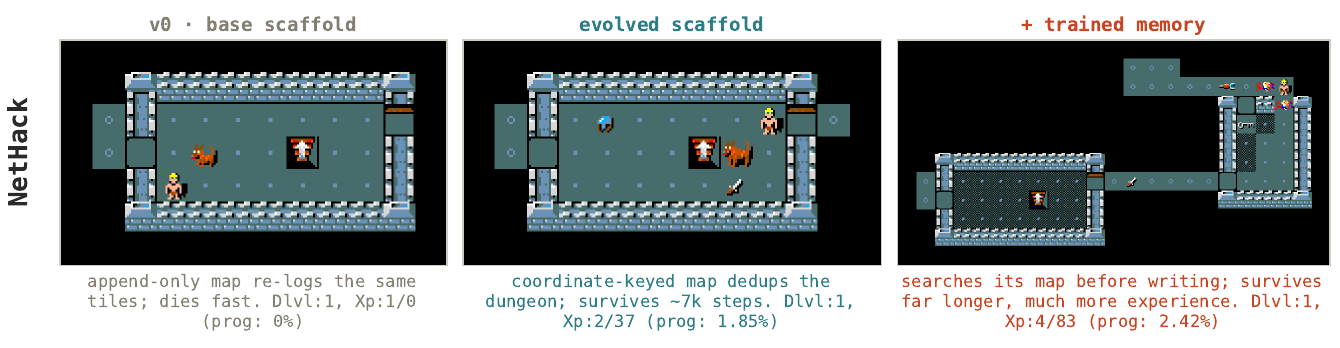}
  \caption{\textbf{Qualitative behavior across the two optimization stages.} Each row is an environment; its three cells show the base scaffold \texttt{v0}, the evolved scaffold, and the $+$ trained specialist on a representative evaluation episode. The note under each cell reports the agent's behavior and its progression rate (\emph{prog.}); for NetHack it also lists the dungeon level (\texttt{Dlvl}, the agent's depth) and the experience level and points (\texttt{Xp}). \textbf{Crafter}: the base agent only gathers wood; the evolved scaffold crafts stone tools, builds a furnace, and mines iron; and the trained specialist reaches that same crafting tier while also feeding itself ($9\to55\to59\%$ progression). \textbf{MiniHack} Corridor-R3, a task that requires navigating branching corridors to reach a goal staircase: neither the base nor the evolved scaffold reaches the staircase, whereas the trained specialist solves the task ($0\to0\to100\%$). \textbf{NetHack}: the base agent dies at experience level~1 within a few hundred steps; the evolved scaffold survives thousands of steps and reaches experience level~2; and the trained specialist survives far longer and reaches experience level~4 ($0\to1.85\to2.42\%$ progression).}
  \label{fig:qualitative}
\end{figure}

\paragraph{Qualitative examples.}
Figure~\ref{fig:qualitative} traces, for each environment, a representative episode at each optimization stage, illustrating memory's effect on behavior.
In Crafter, the base \texttt{v0} agent loops on basic survival. The evolved scaffold organizes its map and inventory well enough to craft stone tools and a furnace and to mine iron, and the trained specialist reaches the same crafting tier while also sustaining itself with food.
In MiniHack's Corridor-R3, a task that requires navigating branching corridors to reach a goal staircase, both the base and the evolved scaffold exhaust the step budget without reaching it, whereas the trained specialist threads the corridors and solves the task.
NetHack is the sharpest illustration of the delayed-consequence problem. With an append-only map, the base agent re-logs redundant observations and never builds on them, dying at experience level~1 within a few hundred steps. The evolved scaffold's coordinate-keyed map (Figure~\ref{fig:scaffold-evolution}) lets it survive thousands of steps and reach experience level~2. The trained specialist, which searches its map before writing rather than logging blindly, survives far longer still and climbs to experience level~4.
These traces make concrete what the aggregate metrics summarize: improving memory management alone contributes significantly to reshaping the agent's task behavior.

\section{Related Work} \label{sec:related}

\paragraph{External memory for language models.}
The fixed-size context window bounds an LLM's working-memory buffer, and motivates a range of external memory designs.
Retrieval-augmented generation~\citep{lewis2020rag} couples a retriever with a language model, fetching relevant passages from a document store at inference time;
MemGPT~\citep{packer2023memgpt} pages information in and out of context using OS-inspired memory management;
Generative Agents~\citep{park2023generative} maintain a timestamped memory stream from which they retrieve, reflect, and plan;
A-MEM~\citep{xu2025amem} equips LLMs with active decisions on what to retain and forget;
MemoryBank~\citep{zhong2024memorybank} maintains a persistent long-term store across sessions.
MemLLM~\citep{modarressi2024memllm} trains models to interact with a dedicated read-write memory module, and Self-Notes~\citep{lanchantin2023learning} trains models to interleave reasoning with memory tokens---both closest to our \textit{proficiency} axis.
More recent works promote memory operations to \textbf{learnable policy actions} and train models to decide when to store, retrieve, or discard~\citep{yu2026agemem,zhou2025mem1,zhang2026memrl,yuan2025memsearcher,yan2025memoryr1}.
MemAct~\citep{zhang2025memact} shares our ``memory as action'' insight but operates on the context window itself rather than on external files.
MemSkill~\citep{zhang2026memskill} also frames memory operations as evolvable skills whose repertoire grows by reviewing failure cases.
MeMo~\citep{quek2026memo} trains a separate, dedicated memory model that a frozen base LLM queries at inference---an architecture close to our trained memory specialist deployed alongside a frozen task model---but its memory model encodes static document knowledge for question answering rather than the memory-management decisions of a long-horizon agent.
Our work differs in coverage: we optimize both the memory \textit{structure} (the scaffold) and the model's parametric \textit{proficiency} at using it, whereas prior work typically addresses only one axis. 
 
\paragraph{Cognitive science of memory management.}
Our work draws on \textit{metamemory}---the capability to monitor and regulate one's own memory processes---introduced by \citet{flavell1979metacognition} and formalized by \citet{nelson1990metamemory} as a monitor--control loop between a meta-level and an object-level. 
The Extended-Mind thesis~\citep{clark1998extended} argues that such aids---notes, diagrams, files---are part of the cognitive system, not mere accessories.
These cognitive-science perspectives have recently been applied to LLMs: 
CoALA~\citep{sumers2023cognitive} explicitly separates working- from long-term-memory in a language-model architecture;
and MetaMem~\citep{xin2026metamem} introduces a meta-memory layer that self-improves its knowledge-retrieval strategies across tasks.
\methodname{} turns the \textit{metamemory} concept into a \textbf{concrete optimization target} rather than an interpretive lens or a symbolic rule set: the scaffold loop shapes the memory structure, and the training loop sharpens the model's proficiency at the ``monitor--control'' decision-making process.
 
\paragraph{Automated agent optimization.}
A growing line of work automates the optimization of LLM agent systems.
The Automated Design of Agentic Systems framework (ADAS)~\citep{hu2024automated} searches over agent architectures in code;
AFlow~\citep{zhang2024aflow} casts agentic workflow generation as Monte Carlo tree search over code-represented workflows;
DSPy~\citep{khattab2023dspy} compiles LM pipelines into optimized prompt chains;
PromptBreeder~\citep{fernando2023promptbreeder} and APE~\citep{zhou2023language} evolve or search over prompts.
Closest to our setting, MemEvolve~\citep{zhang2025memevolve} evolves memory architectures from a modular design space of ``encode, store, retrieve, and manage'' components; and EvolveMem~\citep{liu2026evolvemem} casts memory optimization as an ``AutoResearch'' loop in which an LLM diagnoses per-question failure logs and proposes retrieval-configuration changes.
Our scaffold loop (Outer-loop~\#1) shares a similar ``diagnose-and-revise'' structure but rewrites the agent's code, prompts, and memory-file schema rather than tuning a fixed retrieval configuration, and draws its update signal from \textbf{complete long-horizon trajectory analysis} (up to $10^5$ steps) rather than from per-question logs in multi-session QA.
These methods are also inference-only, whereas \methodname{} additionally trains the model's memory proficiency (Outer-loop~\#2).
 
\paragraph{LLMs in games.}
Long-horizon game environments have become a standard testbed for LLM agents.
Voyager~\citep{wang2023voyager} uses LLMs to acquire skills and maintain a reusable skill library in Minecraft; 
DEPS~\citep{wang2023describe} applies structured ``describe--explain--plan--select prompting'' to open-world Minecraft tasks;
NetPlay~\citep{jeurissen2024netplay} is the first zero-shot LLM agent applied to full NetHack~\citep{kuttler2020nethack}, underscoring the difficulty of dynamic context management at that scale.
In the embodied-agent paradigm more broadly, ReAct~\citep{yao2022react} interleaves reasoning with acting;
Reflexion~\citep{shinn2023reflexion} draws on verbal self-reflection across episodes;
Inner Monologue~\citep{huang2022inner} grounds LM planning in environment feedback;
and ExpeL~\citep{zhao2024expel} extracts reusable insights from past trajectories to guide future decisions without parameter updates.
We use game environments for their \textbf{stochastic} nature and \textbf{long horizons}, but study a different perspective: memory management as the primary lever, rather than reasoning, planning, or retrieval architectures.

\section{Conclusion} \label{sec:conclusion}

This work demonstrates that memory management is an independently learnable, high-leverage skill for LLM agents. With \methodname{}, we factored this skill into two axes---\textit{structure} and \textit{proficiency}---and automated their optimization through meta-LLM-driven outer loops. 
Targeting memory alone, without modifying the gameplay model's weights, improved performance ${\sim}2\times$--$4\times$ across three long-horizon game environments, approaching the level of frontier proprietary systems.
These findings also validate the \textit{metamemory} perspective from cognitive science as a productive framework for LLM agent design: skilled memory use is learned through practice and feedback, not designed into fixed architectures. The file-system interface, the scaffold revisions, and the trained memory specialist are all instances of this principle---the model acquires memory expertise by having its memory decisions observed, evaluated, and improved.
A further finding is that long-horizon task improvement can decompose into trajectory-level review and targeted revision---a workflow that meta-LLMs can execute autonomously where human review of full episode traces is intractable. Applying this decomposition to other agent capabilities beyond memory is a promising direction.

\section{Limitations and Future Work} \label{sec:limitations}

Our current study has several limitations, each suggesting a direction for further work. First, the memory we study is \textit{episodic}: the file system starts fresh at the beginning of each episode, and a natural extension is a \textit{persistent} memory that carries knowledge across episodes. Second, our experiments are on game environments, which are well-suited to studying memory---long horizons, procedural generation, and rich information-management demands---and the approach could also be applied to real-world, memory-intensive tasks. Third, since the three games differ in structure and objectives, we optimize a separate scaffold and memory specialist for each; whether a single scaffold or specialist can be shared across environments remains to be explored.

\section{Broader Impacts} \label{sec:broader-impacts}

\methodname{} shows that automated scaffold optimization and targeted training of memory management as a distinct capability effectively improves performance on long-horizon tasks. This also lowers the model-scale threshold at which long-horizon agents become practical, an accessibility gain for open-weight deployment. The same techniques could be adapted to tasks beyond games. The released artifacts are not directly applicable to high-stakes deployment without further safety review.

\begin{ack}
We thank Kaiyue Wen, James Burgess, Laura Bravo-S\'anchez, and Mark Endo for their insights and helpful discussions.
\end{ack}

\bibliographystyle{plainnat}
\bibliography{references}

\appendix

\section{Implementation Details} \label{app:setup}

This appendix gives implementation configuration organized by components. The complete prompt templates for both outer loops, together with all code, are released in our codebase: \url{https://github.com/autoLearnMem/AutoMem}.

\subsection{Game environment configuration} \label{app:env-config}

\paragraph{BALROG environments.} 
We use the BALROG harness as released~\citep{paglieri2024balrog}, with minor configuration changes: 
\emph{(i)} We set \texttt{autopickup=True} on MiniHack to align with the NetHack configuration. 
\emph{(ii)} The default fallback action when the LLM outputs no parseable action string is set to \texttt{search} where the task's action space includes it (NetHack and a subset of MiniHack tasks) instead of the BALROG default \texttt{esc}, and kept as \texttt{Noop} for Crafter. 
All baseline runs with \qwen{} use the same updated settings as our \methodname{} method. 

Crafter is configured with a 64$\times$64 world area, a 9$\times$9 agent view, dense reward, \texttt{unique\_items=True}, and \texttt{max\_episode\_steps=2000}. 
MiniHack covers eight tasks: Boxoban-Hard, Boxoban-Medium, MazeWalk-9$\times$9, MazeWalk-15$\times$15, Corridor-R3, CorridorBattle-Dark, Quest-Easy, and Quest-Medium, with \texttt{max\_episode\_steps=100}, \texttt{penalty\_step=$-0.01$}. 
NetHack uses \texttt{max\_episode\_steps=100,000} and \texttt{no\_progress\_timeout=150}. 
For all three environments, \texttt{skip\_more=True} suppresses the \texttt{--More--} prompt.

\paragraph{Evaluation seeds.} 
We evaluate \methodname{} using a fixed list of 10 seeds, $[42, 43, \ldots, 51]$, paired with per-environment episode counts: 10 episodes for Crafter, 5 episodes per task for MiniHack ($5 \times 8 = 40$ episodes), and 5 episodes for NetHack. 
Episode index $i$ always uses seed $42+i$, so per-episode results are directly comparable across scaffold \& memory-trained versions.

\subsection{Outer-loops and training pipeline} \label{app:method-impl}

\paragraph{Outer-loop \#1.} 
The scaffold optimization is driven by Claude Opus~4.6 with \texttt{--effort max}. A revision is accepted only if its average progression on the same fixed eval seeds strictly exceeds the previous iteration's. When a fresh revision fails the gate, up to 1 retry is run within the same session where the meta-LLM is given the failed evaluation log and asked to revise again. If that also fails, the meta-loop is restarted from a clean session.

\paragraph{Outer-loop \#2.} 
The pipeline runs in three stages: (a)~training-data collection, (b)~data-engine selection, and (c)~LoRA finetuning followed by two-model inference deployment. Stages (a) and (b) operate on the final scaffold from Outer-loop~\#1 (V5 for Crafter, V4 for MiniHack, V2 for NetHack); the trained adapter is then deployed back into that same scaffold under the two-model inference architecture.

\textbf{Stage (a)~training-data collection}: The base model plays 100 episodes for Crafter, 50 episodes per task for MiniHack (50 $\times$ 8 = 400 episodes total), and 50 episodes for NetHack under the final scaffold. Seeds are drawn randomly and explicitly disjoint from the 10 evaluation seeds $[42, \ldots, 51]$, so no train/eval seed contamination is possible.

\textbf{Stage (b)~training engine:} The training engine is driven by Claude Opus~4.7 with \texttt{--effort max}. Each iteration it (i)~sets the dataset composition (\emph{e.g.,} which episodes to draw from, which standards to apply, minimum dataset size); (ii)~selects training examples from the trace pool; and (iii)~picks a LoRA training configuration. The trained model is then evaluated and the meta-LLM uses the eval trajectories to refine its choices on the next pass.
Advisory per-environment configurations are given at the beginning of the loop as a starting prior. The training engine is free to deviate from these based on its own analysis.

After the meta-LLM produces the selection, a deterministic postprocessing step is applied: format artifacts (code-block wrappers) are cleaned from assistant messages; examples containing only gameplay action commitment with no memory operations are filtered out, since they carry no memory-op training signal for the specialist; and for examples that contain both memory operations and an action commitment, the gameplay action part is trimmed, retaining only the memory-operation reasoning.
The final training sets selected by the training engine contain 1597 examples for Crafter, 444 for MiniHack, and 800 for NetHack.

\textbf{Stage (c)~LoRA training.} We use \texttt{cutoff\_len=16384}, bf16 precision, AdamW, and a cosine learning-rate schedule with \texttt{warmup\_ratio=0.05}. We run training on two GPUs using DeepSpeed ZeRO-3. The LoRA hyperparameters differ across environments, and we report the per-environment configuration used to produce the deployed adapter:
\begin{itemize}
  \item \textbf{Crafter:} \texttt{lora\_rank=256}, \texttt{lora\_alpha=512}, \texttt{lora\_dropout=0.0}, \texttt{effective\_batch\_size=32}, \texttt{learning\_rate=5e-5}, \texttt{num\_train\_epochs=4}, \texttt{target\_modules=attention-only}.
  \item \textbf{MiniHack:} \texttt{lora\_rank=128}, \texttt{lora\_alpha=256}, \texttt{lora\_dropout=0.0}, \texttt{effective\_batch\_size=16}, \texttt{learning\_rate=5e-5}, \texttt{num\_train\_epochs=3}.
  \item \textbf{NetHack:} \texttt{lora\_rank=256}, \texttt{lora\_alpha=512}, \texttt{lora\_dropout=0.0}, \texttt{effective\_batch\_size=32}, \texttt{learning\_rate=5e-5}, \texttt{num\_train\_epochs=1}.
\end{itemize}

\section{Memory-scaffold evolution by iteration} \label{app:scaffold-evolution}

This appendix lists the main changes Outer-loop~\#1 made to the agent scaffold at each iteration, for all three environments. Every change edits the agent's prompts, its memory-file layout, or the automatic hints the scaffold computes from the observation and shows the agent; the two-phase \textsc{LOG}/\textsc{PLAN} loop and the memory-operation interface are otherwise unchanged.

\paragraph{Crafter (v0\,$\to$\,v5).}
\begin{itemize}
  \item \textbf{v1:} Pre-load \texttt{game\_knowledge.txt} with the Crafter crafting tree and survival/placement rules, and replace the empty \texttt{strategy.txt} with a \texttt{goals.txt} template that the agent keeps up to date in every planning prompt. Warn automatically when health, food, or drink runs low, or when an action keeps failing.
  \item \textbf{v2:} State the inventory change since the previous step in the \textsc{LOG} prompt (\emph{e.g.,} ``gained +1 wood''). Add a 22-item achievement checklist (\texttt{progress.txt}), and detect back-and-forth movement loops.
  \item \textbf{v3:} For crafting near a table: escalate the warning as a craft keeps failing; after a table or furnace is placed, list what can now be built; when a table is in view, list the specific items the current inventory has the materials to make.
  \item \textbf{v4:} From the inventory, estimate the next crafting step and indicate which resource to gather next. Warn when health drops or a hostile is nearby, and list untried one-step achievements.
  \item \textbf{v5:} Before a craft or place action is committed, verify the agent has the required materials and otherwise block the action, so it stops attempting impossible crafts. Also block a sleep action next to a monster, reject a \texttt{Noop}-action once it has repeated several times, and warn when the inventory has not grown for many steps.
\end{itemize}

\paragraph{MiniHack (v0\,$\to$\,v4).}
\begin{itemize}
  \item \textbf{v1:} Track the agent's recent actions and warn it when it paces back and forth or repeats the same move. Provide per-task rules (\emph{e.g.,} a dedicated rule sheet for the Boxoban push puzzles).
  \item \textbf{v2:} Record which map cells the agent has already visited and report which neighbouring directions are still unexplored. When the game asks a follow-up question (\emph{e.g.,} ``In what direction?''), prepend a directive to reply with a single direction word. Also auto-load the agent's most recent action-log entries into its prompt.
  \item \textbf{v3:} When the goal staircase is visible on the map, add a planning-prompt directive toward it. When the agent revisits the same cell, add a directive toward an unexplored passable direction. When the game asks a direction question, replace the long planning prompt with a short direction-only one.
  \item \textbf{v4:} Fix the map-reading the hints above depend on: parse passable directions correctly, read tiles per task (\emph{e.g.,} the \texttt{\#} symbol is an impassable bar in Boxoban), switch to another direction when a directed move leaves the position unchanged (\emph{e.g.,} hits a wall), route around lava, and flag nearby doors.
\end{itemize}

\paragraph{NetHack (v0\,$\to$\,v2).}
\begin{itemize}
  \item \textbf{v1:} Add the \texttt{<|UPSERT\_MAP|>} operation, which keeps one entry per map coordinate so a new sighting of a tile overwrites the old entry instead of piling up duplicates. Automatically trim the action log, pre-fill \texttt{strategy.txt} with the main goal and safety rules, and expand the game rules (\emph{e.g.,} avoiding unsafe corpses).
  \item \textbf{v2:} When the agent is on or next to a downward staircase, issue a high-priority directive to descend. Maintain two memory files from the observation every step, a \texttt{current\_status.txt} and an up-to-date \texttt{inventory.txt}. Warn on back-and-forth movement.
\end{itemize}

\end{document}